\documentclass[letterpaper, 10 pt, conference]{ieeeconf}  

\usepackage{graphicx}
\usepackage{xspace} 
\usepackage{algorithm}
\usepackage{algpseudocode}
\usepackage{multirow}
\usepackage{ctable}
\usepackage{tabularx}
\usepackage{booktabs}
\usepackage{amsmath}
\usepackage{pifont}
\usepackage{hyperref}
\usepackage[backend=bibtex,style=ieee,natbib=true]{biblatex}

\usepackage{amsmath,amsfonts,bm}

\def\eqref#1{equation~\ref{#1}}

\def\1{\bm{1}}

\DeclareMathAlphabet{\mathsfit}{\encodingdefault}{\sfdefault}{m}{sl}
\SetMathAlphabet{\mathsfit}{bold}{\encodingdefault}{\sfdefault}{bx}{n}

\IEEEoverridecommandlockouts                              

\title{\LARGE \bf
Retrieve-Augmented Generation for Speeding up Diffusion Policy without Additional Training
}

\author{$^{1}$Sodtavilan Odonchimed, $^{1}$Tatsuya Matsushima, $^{1}$Simon Holk, $^{1}$Yusuke Iwasawa, $^{1}$Yutaka Matsuo
\thanks{$^{1}$Sodtavilan Odonchimed, Tatsuya Matsushima, Simon Holk, Yusuke Iwasawa, Yutaka Matsuo are with School of Engineering, 
        The University of Tokyo, Japan
        {\tt\small sodoo@weblab.t.u-tokyo.ac.jp}}%
}

\begin{document}

\maketitle
\thispagestyle{empty}
\pagestyle{empty}

\begin{abstract}

Diffusion Policies (DPs) have attracted attention for their ability to achieve significant accuracy improvements in various imitation learning tasks. However, DPs depend on Diffusion Models, which require multiple noise removal steps to generate a single action, resulting in long generation times. To solve this problem, knowledge distillation-based methods such as Consistency Policy (CP) have been proposed. However, these methods require a significant amount of training time, especially for difficult tasks.
In this study, we propose RAGDP (Retrieve-Augmented Generation for Diffusion Policies) as a novel framework that eliminates the need for additional training using a knowledge base to expedite the inference of pre-trained DPs. In concrete, RAGDP encodes observation-action pairs through the DP encoder to construct a vector database of expert demonstrations. During inference, the current observation is embedded, and the most similar expert action is extracted. This extracted action is combined with an intermediate noise removal step to reduce the number of steps required compared to the original diffusion step.
We show that by using RAGDP with the base model and existing acceleration methods, we improve the accuracy and speed trade-off with no additional training. Even when accelerating the models 20 times, RAGDP maintains an advantage in accuracy, with a 7\% increase over distillation models such as CP.

\end{abstract}

\section{Introduction}
In the effort to teach behaviors to intelligent agents, imitation learning has been utilized to solve various tasks~\citep{SCHAAL1999233, Osa_2018}. With the success of Diffusion Models in other fields, researchers have been experimenting with these models for imitation learning, yielding great results~\citep{octo, diffusionpolicy, 3d-dp, goal-based-dp, diff-augment-bc}. Among these, Diffusion Policy (DP) ~\citep{diffusionpolicy} has achieved state-of-the-art performance in Behavior Cloning.

Despite recent advances, Diffusion Policies remain computationally expensive because they depend on Diffusion Models. The fundamental issue is that sequential denoising of full Gaussian noise is required to generate a single sample, which significantly increases inference time. 
For example, Diffusion Policy operates using Denoising Diffusion Probabilistic Models (DDPM)~\citep{ddpm}, requiring approximately 100 denoising iterations to generate actions from Gaussian noise, which takes 0.110 seconds to generate one action ~\cite{stanford-cp}, making it difficult to achieve smooth motion.
Reducing the number of denoising steps can speed up inference, but it risks lowering accuracy since the noise may not be fully removed. Therefore, accelerating the inference time of Diffusion Policies is essential. In practice, Diffusion Models often employ efficient sampling techniques, such as DDIM ~\cite{ddim} and DPM++ ~\cite{dpmpp}, which enable generating samples with fewer steps than required during training. These methods tend to suffer from a loss in accuracy as the number of steps is reduced.

Another approach for acceleration is to use Consistency Models (CM) ~\citep{cm, ctm}, which rely on knowledge distillation ~\citep{progdist}. In imitation learning, an example of this is the Consistency Policy (CP) ~\citep{stanford-cp}. However, knowledge distillation methods tend to require significantly more training time as task complexity increases, resulting in additional training costs. Moreover, even small decreases in accuracy can cause covariate shifts that lead to sub-optimal policies, potentially compounding errors in imitation learning ~\citep{ross2011reduction, rajaraman2020toward}.

\begin{figure}[!t]
    \centering
    \includegraphics[width=0.5\textwidth]{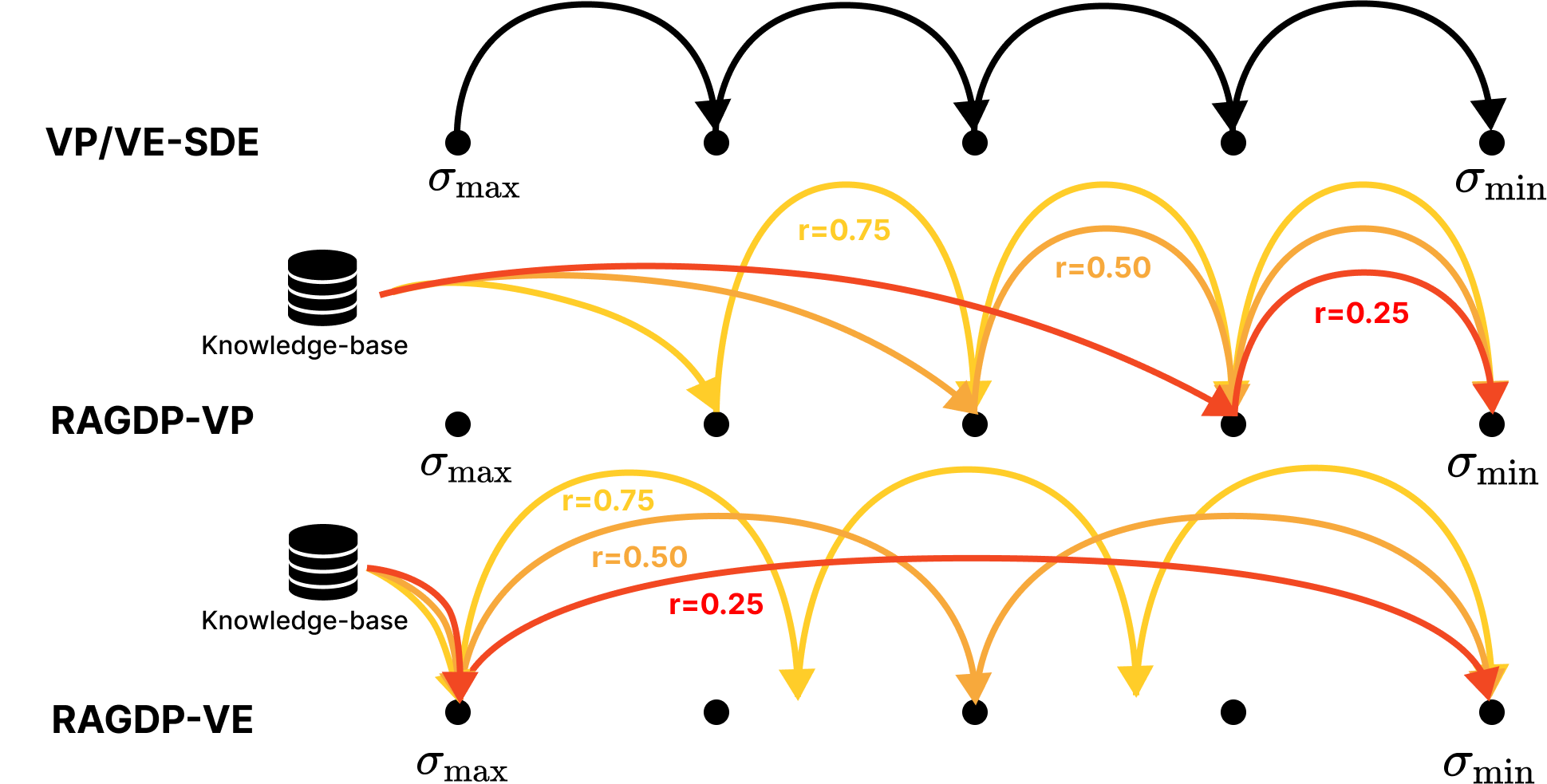}
    \vskip -0.1in
    \caption{\textbf{Diffusion-based Policies and RAGDP} The Diffusion Models generate samples by removing noise from $\sigma_{\mathrm{max}}$ to $\sigma_{\mathrm{min}}$ over $T$ steps. There are two types of noise removal methods: VP-SDE and VE-SDE. RAGDP adapts to each Diffusion Model and can generate actions using two methods (RAGDP-VP and RAGDP-VE). The proposed method can be accelerated by obtaining neighboring values from the knowledge base and reducing the number of noise removal steps by $rT$.}
    \label{fig:cover}
    \vskip -0.3in
\end{figure}
In this study, we propose Retrieve-Augmented Generation for Diffusion Policies (RAGDP) as a new framework for accelerating pre-trained Diffusion Policies without further training. Inspired by Retrieval-Augmented Generation (RAG) ~\citep{rag} in Large Language Models, RAGDP encodes observation-action pairs through the Diffusion Policy encoder and constructs a vector database of expert demonstrations. At inference time, it embeds the current observation and extracts the most similar expert action. This extracted action is combined with Gaussian noise scaled by learned schedule parameters and an intermediate denoising step corresponding to the leap-ratio $r$, reducing the original $T$ diffusion steps down to $rT$ (\autoref{fig:cover}). 
RAGDP enables the acceleration of trained models by performing searches for each inference. It is also adaptable to both VP-SDE and VE-SDE ~\cite{sgm} Diffusion Models.

The proposed methods demonstrate an increase in performance when it comes to the trade-off between speed and accuracy when compared to other methods.
Comparing RAGDP-VP with DPM++ when using a DDPM-based DP, our method achieves a higher task completion rate when accelerating inference by an order of magnitude. This difference is further evident when the model is accelerated 20 times, where the baseline maintains 16\% of the original accuracy, and RAGDP maintains 64\%.
When comparing CP and DPM++ using an EDM-based DP, with RAGDP-VE, our method maintains an advantage over the accuracy and speed trade-off when accelerating 4 times up to 20 times. RAGDP-VE maintains the accuracy of the original model when sped up 4 times and shows a 12\% improvement over CP when sped up 20 times despite not requiring model distillation.
These results demonstrate that RAGDP provides a practical, training-free approach for adapting pre-trained Diffusion Policies to real-time robot control.

\section{Related Work}
\subsection{Diffusion Models for Robotics}


Diffusion models generate data that follows a specific distribution through an iterative denoising process. These models have demonstrated remarkable generative capabilities in the generation of unconditional and conditional images, videos, and 3D objects ~\cite{9878449, cascade, tunefree}.

In recent years, the use of diffusion models has been on the rise in the field of robotics. These models are being applied to various tasks such as immitation learning and task planning, with Diffuser \cite{diffusers} as a typical example. Diffuser performs task planning by simultaneously generating the state of the environment and the trajectory of actions. This led to them achieving higher accuracy than other methods in long-horizon tasks.

In imitation learning, algorithms such as Behavior Cloning ~\cite{bc}, Energy-Based Models ~\cite{ibc}, and Behavior Transformation ~\cite{bet} are commonly used. Still, the emergence of Diffusion Model-based policies has shown promising results with superior performance in tasks requiring multi-step decision-making and robust generalization capabilities. In DP, actions are learned by removing noise from input data, and progressively predicting actions through multiple denoising steps. However, Diffusion Models have the drawback of being slow in real-world applications because they perform noise removal across multiple steps. Therefore, our proposed method builds upon DP and provides a way to accelerate DP by reducing the number of steps needed at inference.

\subsection{Fast sampling methods and their applications to Diffusion Policy}
Diffusion models involve a trade-off between accuracy and the number of sampling steps, creating a need for faster sampling in many applications. There are multiple approaches to accelerate diffusion models: First, knowledge distillation trains a student model to mimic a teacher model with many sampling steps and high generative quality. By learning from the teacher, the student can generate samples with fewer steps, achieving faster inference. Prior work in this area includes CM and Consistency Trajectory Models (CTM) ~\cite{cm, ctm}, as well as Progressive Distillation ~\cite{progdist}. In the context of DP, methods like CP ~\cite{cp} and One-Step Diffusion Policy ~\cite{onestep-dp} apply these ideas to accelerate action generation.

The second method is to use a differential equation solution algorithm. It has been shown that Diffusion Models can be generalized as Score-based Generative Models (SGMs) using stochastic differential equations ~\cite{yang2023diffusion, sgm}. Therefore, the sampling process of Diffusion Models corresponds to the task of solving differential equations, and efficient calculation algorithms are directly related to speed improvement. For the DDPM model, fast sampling methods such as DDIM~\cite{ddim} have been proposed, while other trained models include DPM~\cite{dpm} and DPM++~\cite{dpmpp}. These sampling algorithms can also be applied directly to DP.

The third method is to train Diffusion Models with latent variables. The use of latent variables can reduce the dimensionality of the data to be computed, thus speeding up the process. However, additional model training is required because VAE must be used to convert the latent variables. A typical example is Stable Diffusion~\cite{stable-diffusion}, which uses VQ-VAE~\cite{vqvae} and Latent Diffusion Models. LDP \cite{ldp} is an example of its application in robotics.
Our proposed method builds on the SGM perspective and provides a way to accelerate action generation during inference without any additional training. Moreover, our approach can be combined with existing acceleration techniques to further enhance their performance.

\subsection{Retrieval-Augmented Methods in Robotics}
In robot learning, many methods use retrieval augmentation to enhance performance, typically by improving the quality of training data. For example, Behavior Retrieval methods ~\cite{learning-retrieve, behavior-retrieve} show that selecting high-quality demonstrations allows models to achieve good performance with less data. In addition, DemInf~\cite{mixture-data} proposes a method to acquire and evaluate data quality based on entropy, which improves learning efficiency by prioritizing high-quality samples. These approaches apply retrieval during the creation of learning datasets, while our method focuses on improving inference directly.


Some methods are used for both training and inference; in Motion Planning, the READ ~\cite{r2diff, read_CVPR} method is an example. This method works by retrieving and sorting training data and then retrieving it with initial conditions when planning. Some studies retrieve behavior data by pairing behavior data with another modality; in ReMoDiffuse ~\cite{remo-diff, motion-diff}, behavior generation is conditioned on retrieved samples based on pairs of behavior and instructional text data. These data are intended not only to select training data but also to improve accuracy by retrieving good samples. By contrast, our work applies retrieval only during inference with an already trained model. This approach improves action generation at test time without requiring any additional training. Moreover, the retrieved data can be limited to the dataset already used during training, avoiding the need for any extra dataset processing.

\begin{figure*}[t]
    \begin{center}
    \includegraphics[width=0.95\textwidth]{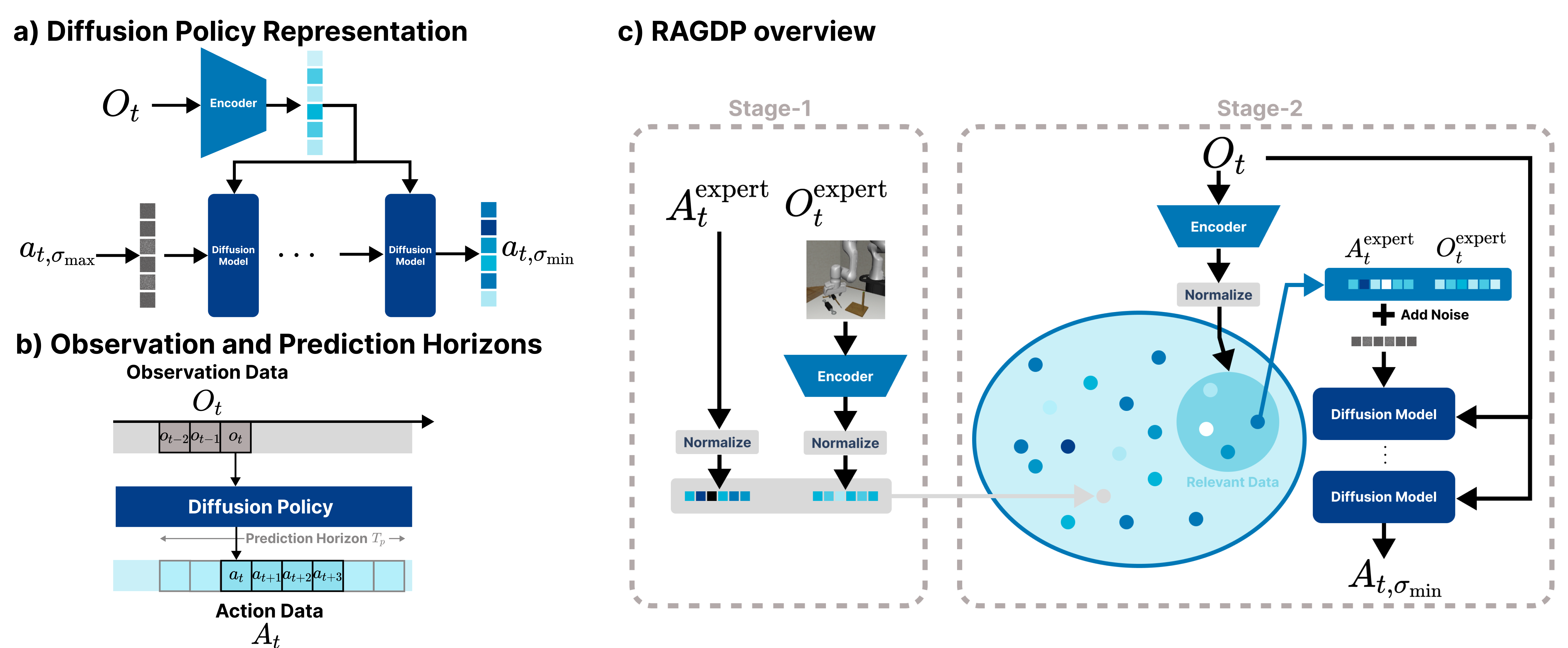}
    \caption{\textbf{a) Diffusion Policy Representation:} Diffusion Policy behaves as a diffusion model that takes data observed from the environment as conditional input and outputs trajectory data. It  generates samples by removing noise from $\sigma_{\mathrm{max}}$ to $\sigma_{\mathrm{min}}$ over $T$ steps. \textbf{b) Observation and Prediction Horizons:} 
The conditional input is $O_t$, chunked by $T_o$ steps of observed data $o_t$, and the generated behavior is $A_t$, chunked by $T_p$ steps of action step $a_t$. \textbf{c) RAGDP overview:} The RAGDP is divided into two parts: the first is the knowledge base part, which is implemented as a vector database of observation and trajectory data pairs; the second is the diffusion model part, which outputs the final trajectory data via Diffusion Policy.
The operation has two steps. Stage-1, which encodes the expert's data into a 1D vector and stores it in a vector database; Stage-2 consists of searching for relevant actions with observations made in the inference environment and generates them using a trained diffusion model.}
    \label{fig:overview}
    \end{center}
    \vskip -0.25in
\end{figure*}

\section{Methodology}
\subsection{Formulation}
We provide the formulation and explain the proposed method in this section. SGMs generalize the Diffusion Models as a stochastic differential equation. Let $t$ denote the time step and $\tau$ denote the diffusion step; let $A_t$ be the trajectory at time step $t$ and $O_t$ be the data of the observed environment at that time. In the Diffusion Models, the direction in which the amount of noise increases is considered the forward process, while the direction in which the amount of noise decreases is considered the reverse process. Let $\sigma(\tau)$ be the sampling scheduler of the Diffusion Model and take the range of $\sigma \in [\sigma_\mathrm{min}, \sigma_\mathrm{max}]$. Then, we define two functions $F: \mathbb{R}^{D_A} \times [\sigma_\mathrm{min}, \sigma_\mathrm{max}] \to \mathbb{R}^{D_A} $ and $G: [\sigma_\mathrm{min}, \sigma_\mathrm{max}] \to \mathbb{R}$. Where $D_A$ is the dimension size of $A_t$. At this point, the forward process is as follows ~\citep{sgm}:
\begin{equation}
\label{eq:diff-forward}
    dA_t (\tau) = F(A_t (\tau), \sigma(\tau)) d\sigma + G(\sigma(\tau)) d\omega.
\end{equation}
Reverse process is as follows:
\begin{align}
\label{eq:diff-backward}\nonumber
    &dA_t (\tau) = [ F(A_t (\tau), \sigma(\tau)) \\  &- \frac{1}{2} G(\sigma(\tau)^2 \nabla_{A_t} \log p_\sigma(A_t (\tau) | O_t) ] d\sigma 
    +G(\sigma(\tau)) d\omega.
\end{align}

\subsubsection{Variance Preserving Stochastic Differential Equations}
In the equation above, when $\sigma(\tau) = \tau$ and the functions are $F(A_t(\tau), \tau) = -\frac{1}{2} \beta (\tau) A_t (\tau)$ and $G(\tau) = \sqrt{\beta(\tau)}$, then the equation represents Variance Preserving Stochastic Differential Equations (VP-SDE). When the two functions are applied to the~\autoref{eq:diff-forward} and the differential equation is solved, the general solution is as follows:
\begin{equation}
    \label{eq:vp-sol}
    A_t(\tau) = \alpha(\tau) A_t + \sigma(\tau) \epsilon \quad \mathrm{where} \quad \epsilon \sim \mathcal{N}(0, \mathbf{I}).
\end{equation}
Where $\alpha(\tau)$ and $\sigma(\tau)$ are functions computed from $\beta(\tau)$ and have properties such as $\alpha(\tau)^2 + \sigma(\tau)^2 = 1$. Therefore, in VP-SDE, noise and data are mixed as a ratio at each step $\tau$, resulting in $\sigma(\tau) \in [0, 1]$. DDPM is classified as this type of Diffusion Models.

\subsubsection{Variance Exploding Stochastic Differential Equations}
Then, if the function is set $F(A_t(\tau), \tau) = 0$ and $G(\tau) = \sqrt{2\sigma(\tau)}$, called Variance Exploding Stochastic Differential Equations (VE-SDE). The general solution in this case is as follows:
\begin{equation}
    \label{eq:ve-sol}
    A_t(\tau) = A_t + \sigma(\tau) \epsilon \quad \mathrm{where} \quad \epsilon \sim \mathcal{N}(0, \mathbf{I}).
\end{equation}
In VE-SDE, there is no limit to the amount of noise, and $\sigma \in [\sigma_\mathrm{min}, \sigma_\mathrm{max}]$. VE-SDE based EDM~\citep{edm} models were employed in our experiments.

Let $o_t$ be the observed data at a certain time and $a_t$ be the behavior taken at that time, and $\mathcal{D} = \{(o_0 ^{(i)}, a_0 ^{(i)}, (o_1^{(i)}, a_1^{(i)}), ..., (o_\mathcal{T}^{(i)}, a_\mathcal{T}^{(i)})\}_{i=1} ^ {N}$ be the training data of the model, where $N$ is the number of episodes collected by the expert. The behavior of the Diffusion Policy is illustrated in~\autoref{fig:overview} a). 

\subsubsection{Training algorithm}
In the Diffusion Policy, the model takes observation data as input and outputs behavioral action data. The input observation data is chunked for the past $T_o$ steps $O_t = [o_t, o_{t-1}, ... ]$. Let $N_\mathrm{data}$ be the number of $O_t$ chunks in all episodes.
The output action data is chunked for $T_p$ steps of action step $a_t$ and is $A_t$. Only $T_a$ steps of it are executed.
To generate $A_t$ using the Diffusion Model,~\autoref{eq:diff-backward} can be utilized. $\nabla_{A_t} \log p_\sigma(A_t (\tau) | O_t)$ in~\autoref{eq:diff-backward} is called the score function and is the quantity that the model should acquire in training $s_\theta = \nabla_{A_t} \log p_\theta(A_{t, \tau} | O_t)$. The optimization algorithm for training is called score matching and is expressed by the following equation:
\begin{eqnarray}
\label{eq:train}
    \mathcal{L}(\theta) = \mathbb{E}\left[ \left| s_\theta(A_{t, \tau}, \sigma_\tau, O_t) - \nabla_{A_t} \log p_{\sigma_\tau}(A_{t, \tau}| O_t) \right|^2\right].
\end{eqnarray}

\subsection{Retrieve-Augmented Generation for Diffusion Policy}
\subsubsection{Retrieval}
The knowledge base of RAGDP is a vector database consisting of pairs of observation data and corresponding expert trajectory data. The key vector at retrieval is the observed data of the training data, and the value vector at retrieval corresponds to the normalized trajectory data $A_t$ of the training data. 
In this study, we introduce $f$ as a function that creates an embedding space. If the observed data $O_t$ is a state, $f$ becomes a simple normalization function $f(O_t) = (O_t - O_{\mathrm{avg}})/ O_{\mathrm{std}}$. Here, $O_{\mathrm{avg}}$ and $O_{\mathrm{std}}$ represent the mean and standard deviation of the observed data in the dataset $\mathcal{D}$. If $O_t$ is an image, it becomes an encoder function trained with DP. This is to enable immediate implementation without training new VAE or other models. When we trained a simple VAE and attempted a search in the embedding space, there was no change in performance.

The vector database is searched for each inference. It then obtains the top sample with the highest search similarity. The retrieve at step $t$ of inference is expressed as ~\autoref{eq:retrieve} . The action $A^{\mathrm{exp}}_i$ in the database is retrieved from the index $i$ with the closest distance.
\begin{equation}
\label{eq:retrieve}
    i = \underset{j \in \{1, ... N_{\mathrm{data}}\}}{\operatorname{argmin}} || f(O^{\mathrm{exp}} _j) - f(O_t) ||_2 .
\end{equation}
The 'exp' represents the trajectory of the training data. All observation data are pre-encoded and normalized $z^{\mathrm{exp}}_j = f(O^{\mathrm{exp}} _j)$ before being stored in the database. This is shown in \autoref{fig:overview} c), Stage 1.

\subsubsection{Generation}
\label{subsec:mathod-generation}
After retrieved the nearest point, it is necessary to consider how to utilize it in the generation process. There are two possible methods for generating actions from neighboring points in Diffusion Models. The first method involves adding a specific amount of noise to the neighboring point and applying it to the current state. An example of this is SDEdit ~\cite{sdedit}. The second method involves adding new parameters to the conditional input of Diffusion Models. The latter requires additional training or implementation, so our method adopts the former approach. RAGDP-VP is an algorithm that behaves similarly to SDEdit. The other method, RAGDP-VE, is a sampling technique designed to correspond to VE-SDE type Diffusion Models. 

\textbf{RAGDP-VP}\\
In VP-SDE, the parameters in ~\autoref{eq:vp-sol} are constrained by $\alpha(\tau)^2 + \sigma(\tau)^2 = 1$. This means that the magnitude of the noise and the action are determined by a ratio. Therefore, the action retrieved from the database is used to calculate the final output from the ratio of action and noise corresponding to the starting diffusion step $\tau_0$.RAGDP-VP introduces a hyperparameter leap-ratio $r$, which determines the initial position to start the denoising process.
If the number of diffusion steps is $T$ and the step to start generating is $\tau_0$, then $r=\tau_0/T$. Since the parameter takes the range $0<r<1$, the number of steps to generate samples is $(1-r)T$, which enables faster processing. 
For example, when $r=0$, it is a normal Diffusion Policy; when $r=1$, the retrieved trajectory is executed as is.
In principle, RAGDP-VP can be applied to both VP-SDE and VE-SDE Diffusion Models and Consistency Models. The DDPM-based RAGDP-VP is shown in \autoref{alg:ragdp-vp-sampling}.
\begin{algorithm}[tp]
    \caption{RAGDP-VP (DDPM) Sampling Algorithm}
    \label{alg:ragdp-vp-sampling}
    \begin{algorithmic}[1]
    \Require diffuse rate $r$, total denosing steps $T$, denosing scheduler $\sigma_\tau$, total episode steps $\mathcal{T}$, pretrained model parameters $\theta$, vector database $\{(z_i ^{\mathrm{exp}}, A_i ^{\mathrm{exp}}) | i \in \{ 1, 2, \dots, N_\mathrm{data} \} \}$.
    \For{$t=1$ to $\mathcal{T}$}
        \State Observe $O_t$
        \State $z_t = f(O_t)$
        \State $i \gets \underset{n =1...N_\mathrm{data}} {\operatorname{argmin}} \|z_t - z_n ^{\mathrm{exp}} \|$
        \State $A^{\mathrm{ret}} \gets A_i ^{\mathrm{exp}}$
        \State $\tau^* \gets \lfloor (1-r)T\rfloor $ 
        \State $\epsilon \sim \mathcal{N}(0, \mathbf{I})$
        \State $A_{t, \tau^*} \gets \sqrt{\overline{\alpha}_{\tau^*}} A^{\mathrm{ret}} + \sqrt{ 1- \overline{\alpha}_{\tau^*}} \epsilon$
        \For{$\tau = \tau^*$ to $0$}
                \State $\epsilon \sim \mathcal{N}(0, \mathbf{I})$ \textbf{if} $\tau > 0$ \textbf{else} $z = 0$
                \State $A_{t,\tau-1} = \frac{1}{\sqrt{\alpha_{\tau}}} \left( A_{t,\tau} - \frac{1-\alpha_\tau}{\sqrt{1-\overline{\alpha}_{\tau}}} \epsilon_\theta(A_{t,\tau}, \tau, O_t) \right) + \sigma_\tau \epsilon$  
        \EndFor
        \State Execute $A_{t, 0}$
    \EndFor   
    \end{algorithmic}
\end{algorithm}
\begin{algorithm}[tp]
    \caption{RAGDP-VE (EDM) Sampling Algorithm}
    \label{alg:ragdp-ve-sampling}
    \begin{algorithmic}[1]
    \Require diffuse rate $r$, total denosing steps $T$, denosing scheduler $\sigma_\tau$, total episode steps $\mathcal{T}$, pretrained model parameters $\theta$, vector database $\{(z_i ^{\mathrm{exp}}, A_i ^{\mathrm{exp}}) | i \in \{ 1, 2, \dots, N_\mathrm{data} \} \}$. 
    \For{$t=1$ to $\mathcal{T}$}
        \State Observe $O_t$
        \State $z_t = f(O_t)$
        \State $i \gets \underset{n =1...N_\mathrm{data}} {\operatorname{argmin}} \|z_t - z_n ^{\mathrm{exp}} \|$
        \State $A^{\mathrm{ret}} \gets A_i ^{\mathrm{exp}}$
        \State $n \gets  (1-r)T $ 
        \State $\Delta \tau \gets \lfloor\frac{T}{n}\rfloor$
        \State $\epsilon \sim \mathcal{N}(0, \mathbf{I})$
        \State $A_{t, T} \gets A^{\mathrm{ret}} + \sigma_\mathrm{max} \epsilon$
        \State $\tau \gets T$
        \For{$j = 1$ to $n$}
                \State $A_{t,\tau-\Delta \tau} = A_{t,\tau} + (\sigma_\tau^2 -\sigma_{\tau-\Delta \tau}^2) s_\theta(A_{t,\tau}, \sigma_\tau, O_t)$  
                \State $\tau \gets \tau - \Delta \tau$
        \EndFor
        \State Execute $A_{t, 0}$
    \EndFor   
    \end{algorithmic}
\end{algorithm}

\textbf{RAGDP-VE}\\
In the case of VE-SDE, $\alpha(\tau)$  in ~\autoref{eq:vp-sol} which represents the signal strength of the action is fixed by $\alpha(\tau) = 1$ . Therefore, there is no restriction on the mixing ratio between action and noise signals. The action taken from the database adds noise of a magnitude corresponding to the starting diffusion step $\tau_0$, and the output is obtained where this noise becomes smaller.
Therefore, RAGDP-VE always adds $\sigma_\mathrm{max}$ without changing the amount of initial noise and only changes the number of sample steps. Similarly, a hyperparameter leap-ratio $r$ is introduced, which similarly generates samples by calculating $(1-r)T$ steps. 
RAGDP-VE can be applicable only to VE-SDE-based Diffusion Model and Consistency Models. The EDM-based RAGDP-VE is shown in \autoref{alg:ragdp-ve-sampling}. Stage 2 in \autoref{fig:overview} c) corresponds to this content.

\begin{figure*}[t]
    \begin{center}
    \includegraphics[width=0.95\textwidth]{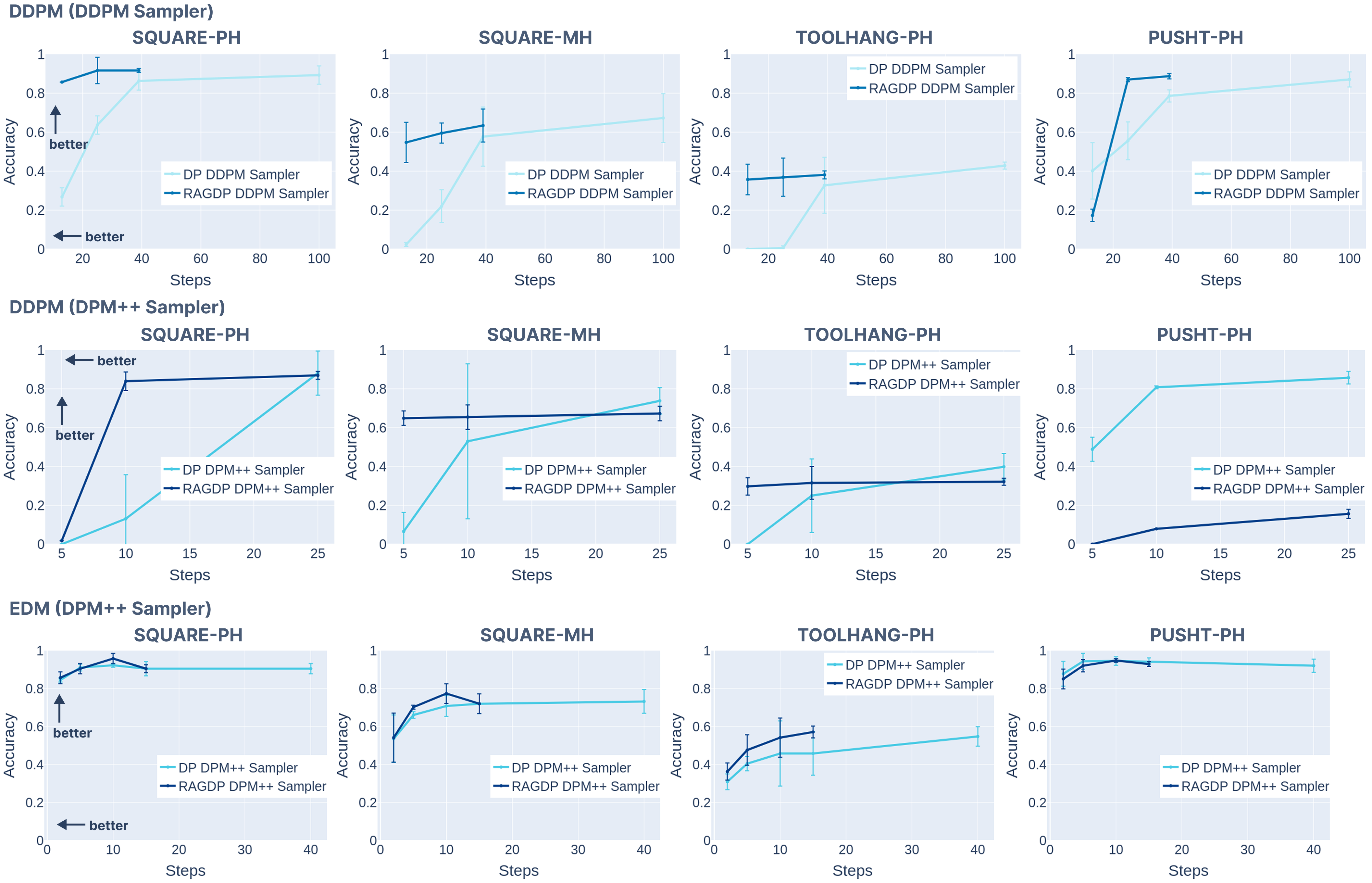}
    \caption{\textbf{The robustness of RAGDP:} The graph displays the robustness when smaller sampling steps are used. We investigate using RAGDP with different base models and samplers. From the top, DDPM model using a DDPM sampler, DDPM model using a DPM++ sampler, and EDM model using a DPM++ sampler. The x-axis of the graph represents the number of sampling steps; the smaller the number, the faster the speed. y-axis represents the accuracy; the larger the accuracy, the better the performance.}
    \label{fig:results-reducing-steps}
    \end{center}
    \vskip -0.25in
\end{figure*}

\section{Experiments}
\subsection{Implementation}
We implemented three diffusion-based control policies for our experiments: a DDPM-driven Diffusion Policy, an EDM-driven Diffusion Policy, and a CTM-driven Consistency Policy.
All three share an identical CNN-UNet backbone with 12 residual blocks: the encoder channels increase from 256 → 512 → 1024, and the decoder performs the process in the reverse order.
We group the models by their stochastic-differential-equation formulation and sampler:
\begin{itemize}
    \item DDPM (VP-SDE). Treated as a variance-preserving diffusion model. It predicts the injected noise ($\epsilon$-prediction) and can be sampled with either the original DDPM solver or the more efficient DPM++ solver.
    \item EDM (VE-SDE). Treated as a variance-exploding diffusion model. It outputs denoised samples directly and is sampled solely with DPM++.
    \item CTM (Consistency Model). Uses its own single-step consistency sampler and is trained by distilling knowledge from the pretrained EDM model.
\end{itemize}
Because our acceleration method is applied only at inference time, each model is trained with its standard procedure—Diffusion Policy training for DDPM and EDM, and Consistency Policy training for CTM. The model details are given in \autoref{tab:models}.
\begin{center}
\begin{table}[htbp]
\caption{\textbf{Details of the model implementation.} Details of the model implementation; the DDPM and EDM models are implemented as DP and the CTM model as CP.}
\centering
\label{tab:models}
\begin{tabular}{ccccc}
\toprule
\multicolumn{1}{c}{Models} & Denoising Steps & Sampler & Prediction Type  \\ \midrule
\multicolumn{1}{c}{DDPM} & 100  & DDPM / DPM++ & epsilon \\ 
\multicolumn{1}{c}{EDM} & 40  & DPM++ & sample \\ 
\multicolumn{1}{c}{CTM} & 4  & CTM & sample \\ 
\bottomrule
\end{tabular}
\end{table}
\end{center}

The vector database is implemented using Facebook AI Similarity Search (FAISS)~\citep{faiss}, which uses a search method that indexes based on L2 distances in Euclidean space. All observation data are pre-encoded and normalized before being stored in the database. Action data are also normalized and stored in the database.

\subsection{Evaluation Setup}
We benchmarked on imitation learning dataset robomimic ~\citep{robomimic}. Square-PH, Square-MH, ToolHang-PH, and PUSHT-PH were employed from robomimic. The details of these tasks are given in \autoref{tab:tasks}. Each task has the following characteristics:
\begin{itemize}
    \item \textbf{SQUARE-PH:} Simple data set, high quality (PH) trajectory data.
    \item \textbf{SQUARE-MH:} a simple data set, but with some low-quality trajectories mixed in.
    \item \textbf{ToolHang-PH:} A complex task dataset with two tools to handle.
    \item \textbf{PUSHT-PH:} A dataset for a task with a small behavioral dimension.
\end{itemize}
The performance of each task is the average of the models trained on 3 different seeds. There were 56 different seeds in the evaluation environment, and a total of 168 accuracy measurements were taken for each task. The evaluation seed was not included in the training seed. For all experiments, robotics state data was used for observations.
The column ``Steps'' in ~\autoref{tab:tasks} specifies the maximum number of steps allowed per episode. Here, SQUARE-PH, MH and PUSH-T are single-step tasks, while TOOLHANG-PH is multi-step task as it moves multiple objects.

To evaluate how well the methods maintain the original performance, we define  ``recovery rate''  as a metric. Recovery rate is calculated as $A_{\mathrm{accelerated}} / A_{\mathrm{base}}$, where $A_{\mathrm{accelerated}}$ represents the accuracy of the acceleration method, and $A_{\mathrm{base}}$ represents the accuracy of the original model.
\begin{center}
\begin{table}[htbp]
\caption{\textbf{Task Details.} \#Rob: number of robots, \#Obj: number of objects, ActD: action dimension, PH: proficient-human demonstration, MH: multi-human demonstration, Steps: max number of rollout steps.}
\centering
\label{tab:tasks}
\begin{tabular}{ccccccccc}
\toprule
\multicolumn{1}{c}{Task} & \#Rob & \#Obj  &  \#ActD  & Demos  &  Steps \\ \midrule
\multicolumn{1}{c}{Square PH} & 1  & 1 &  7  & 200 & 400 \\ 
\multicolumn{1}{c}{Square MH} & 1  & 1 &  7  & 300 & 400 \\ 
\multicolumn{1}{c}{ToolHang PH} & 1  & 2 &  7  & 200 & 700 \\
\multicolumn{1}{c}{Push-T PH} & 1  & 1 &  2  & 200  & 300 \\ 
\bottomrule
\end{tabular}
\end{table}
\end{center}

\subsection{Robustness of RAGDP}
We first perform an extensive evaluation of the trade-off between the number of sampling steps and the accuracy of our method and compare it with baselines. By reducing the number of sampling steps, we can evaluate how this affects the accuracy. We investigate the performance of regular DDPM and the efficient DPM++ sampling method when extended with RAGDP.

The results are shown in \autoref{fig:results-reducing-steps}. Overall, we see a clear improvement in both accuracy and the number of steps needed in most environments. Not surprisingly, using RAGDP with DDPM shows a strong gain in accuracy as the number of steps decreases. More interestingly, even when using the more efficient DPM++ sampler, we see improvement in accuracy. Despite Square-MH having mixed quality demonstration data, we still see an increase over the baseline, which indicates some robustness towards mixed quality data.

To better understand how much accuracy diverges from the DDPM and EDM base models, we compare RAGDP with both DPM++ and CP by measuring the recovery rate. \autoref{tab:recovery-rate}, shows for the DDPM case that RAGDP is able to maintain a higher accuracy when the model is sped up by an order of magnitude. Once inference is accelerated by $\times 20$, RAGDP-VP maintains a 64\% recovery rate compared to 17\% for DPM++. When using EDM, RAGDP-VE is able to maintain and even slightly improve task performance over the original model when having a $\times 4$ speed increase. Furthermore, as the speed is increased by $\times 10$ and $\times 20$, RAGDP-VE is able to maintain an advantage in accuracy over both CP and DPM++.

\subsection{Comparison with other acceleration methods}
\begin{figure}[tbp]
\begin{center}
\includegraphics[scale=0.12]{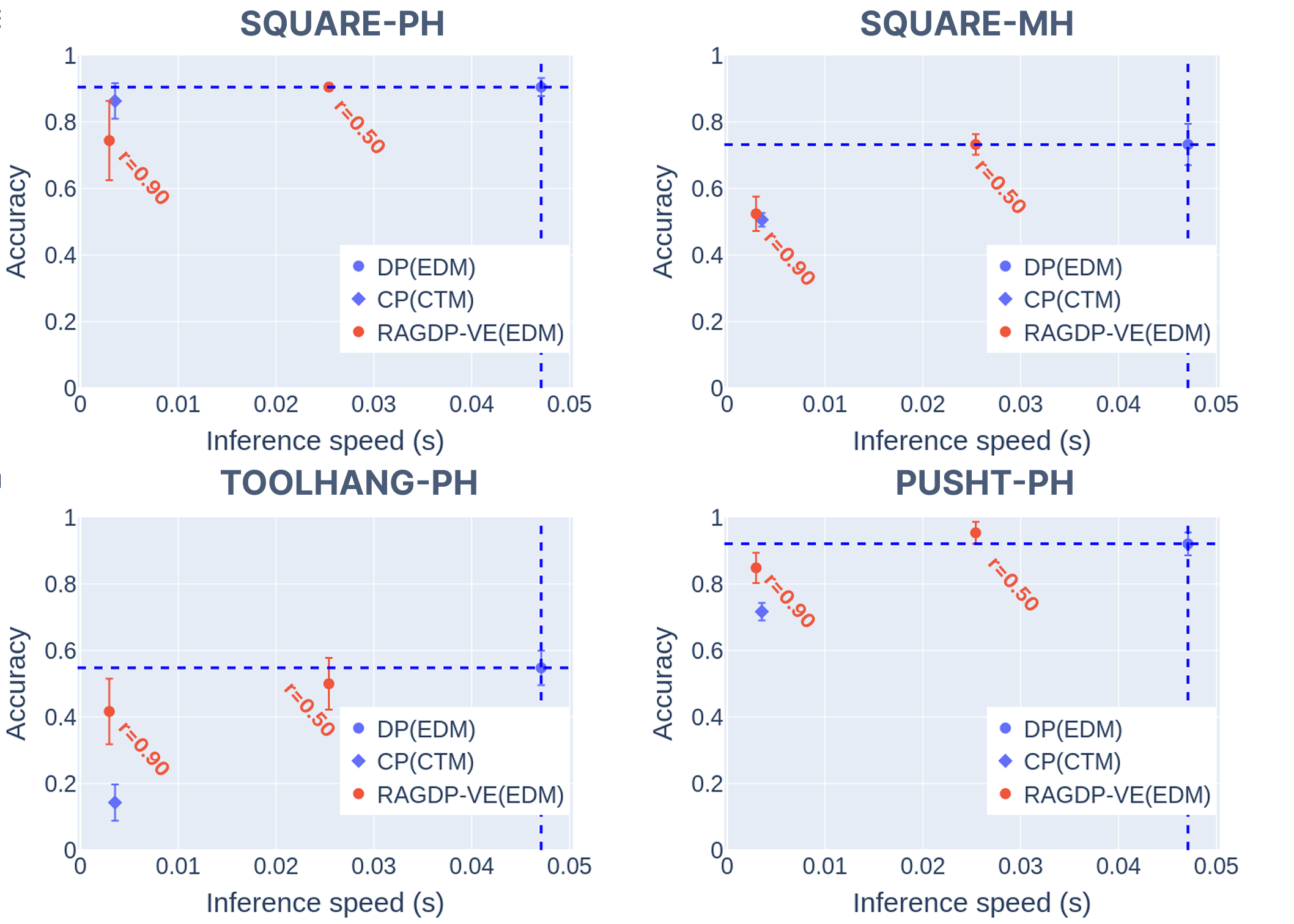}
\end{center}
\caption{\textbf{Inference speed and accuracy:} 3 seeds average rewards are shown as accuracy on the vertical axis and sampling speed on the horizontal axis. Speeds were measured on RTX 3060. The upper left direction of the graph indicates better performance. The results show a comparison of RAGDP-VE with CTM. The results show that by utilising RAGDP-VE for EDM, the accuracy reaches the same or better than that of CTM at the same speed as CTM.}
\label{fig:speeds}
\end{figure}

In this experiment, we compare the accuracy and generation speed of CP and RAGDP-VE. In principle, RAGDP-VE can also be utilized with CP. The EDM-based DP is generated in 40 steps as before, and the number of steps is reduced by changing the hyperparameter $r$. 
The experimental results are shown in \autoref{fig:speeds}, where the lines represent the accuracy and inference speed of the original EDM model. 
We observe that the 4-step CP offers a great reduction in inference time but suffers from lower accuracy compared to EDM-based DP, with significant losses in the harder cases. 
\begin{table}[tbp]
\caption{\textbf{Average performance of task completion.} The results of calculating the average task performance for each method using benchmark tasks. The results of $\times4$, $\times10$, and $\times20$ speedups are shown.}
\centering
\label{tab:cp-ragdp-acc}
\begin{tabular}{lcccc}
\toprule
\multicolumn{1}{l}{Methods} & Base Model & $\times$4 & $\times$10  &  $\times$20 \\ \midrule
\multicolumn{1}{l}{CP} &  EDM & -  & 61.93\% & 57.38\% \\
\multicolumn{1}{l}{RAGDP-VE} &  EDM & 80.54\%  & 75.10\% & 65.32\% \\
\bottomrule
\end{tabular}
\end{table}

We further illustrate the difference in \autoref{tab:cp-ragdp-acc}. When using a $\times20$ speed-up, RAGDP-VE offers an 8\% improvement in accuracy compared to CP. This advantage is even stronger in the $\times10$ case, where it has a 13\% accuracy advantage over CP. Due to our method being agnostic to the underlying implementation of the DP, we can further improve the flexibility of the trade-off between speed and accuracy of existing acceleration methods, as well as the base models.

\begin{center}
\begin{table}[tbp]
\caption{\textbf{Average recovery rate of task completion.} We calculated the average task completion rate of existing DPs to determine how much it would recover when each acceleration method was applied.}
\centering
\label{tab:recovery-rate}
\begin{tabular}{lcccc}
\toprule
\multicolumn{1}{l}{Methods} & Base Model & $\times$4 & $\times$10  &  $\times$20 \\ \midrule
\multicolumn{1}{l}{DPM++} & DDPM & 99.97\%  & 61.14\% &  16.46\%  \\ 
\multicolumn{1}{l}{RAGDP-VP} & DDPM & 72.58\%  & 68.51\% &  63.98\%  \\ 
\midrule
\multicolumn{1}{l}{DPM++} & EDM & 96.29\%  & 91.84\% &  79.62\%   \\ 
\multicolumn{1}{l}{CP} &  EDM & -  & 75.93\% & 69.44\% \\
\multicolumn{1}{l}{RAGDP-VE} &  EDM & \textbf{103.36\%}  & \textbf{95.72\%} & \textbf{81.86\%} \\
\bottomrule
\end{tabular}
\end{table}
\end{center}

\subsection{RAGDP-VP vs. RAGDP-VE}
To compare the differences between the two proposed methods, we evaluate robustness by showing how accuracy changes when the sample size decreases. Therefore, we evaluated the performance of RAGDP-VP and RAGDP-VE with the DPM++ sampler for both trained EDM models.

\begin{figure}[tbp]
\begin{center}
\includegraphics[scale=0.12]{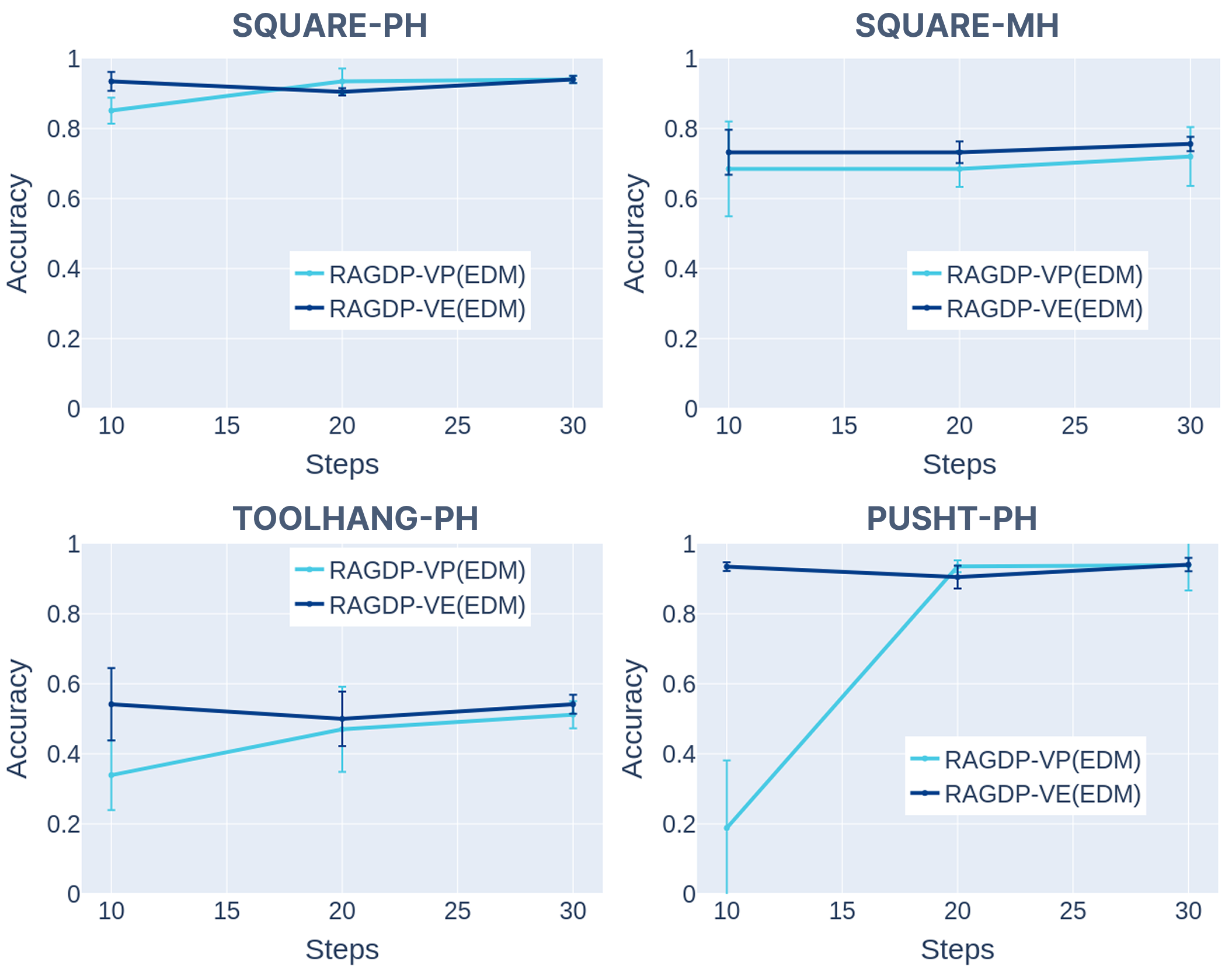}
\end{center}
\caption{\textbf{Comparison of RAGDP-VP and RAGDP-VE performance in VE-SDE based Diffusion Model:} A comparison of VE-SDE-based EDM models in PUSHT-PH and TOOLHANG-PH shows that RAGDP-VP is less accurate with respect to sampling steps, while RAGDP-VE is more robust.}
\label{fig:edm_sde}
\end{figure}

\autoref{fig:edm_sde} shows the results when comparing the change in accuracy for each sampling technique by reducing the number of sampling steps. We can see that both method has a stable accuracy for 20 steps and above while RAGDP-VP performs worse than RAGDP-VE for 10 steps.
RAGDP-VP has a trade-off between faithfulness and realism with respect to neighborhood points.
Therefore, when the parameter $r$ is large, the de-noising step is smaller and the amount of noise given is smaller, so realism tends to be weaker and less accurate. 

\subsection{Choice of the hyperparameter leap-ratio $r$}
\begin{figure}[tbp]

\begin{center}
\includegraphics[scale=0.15]{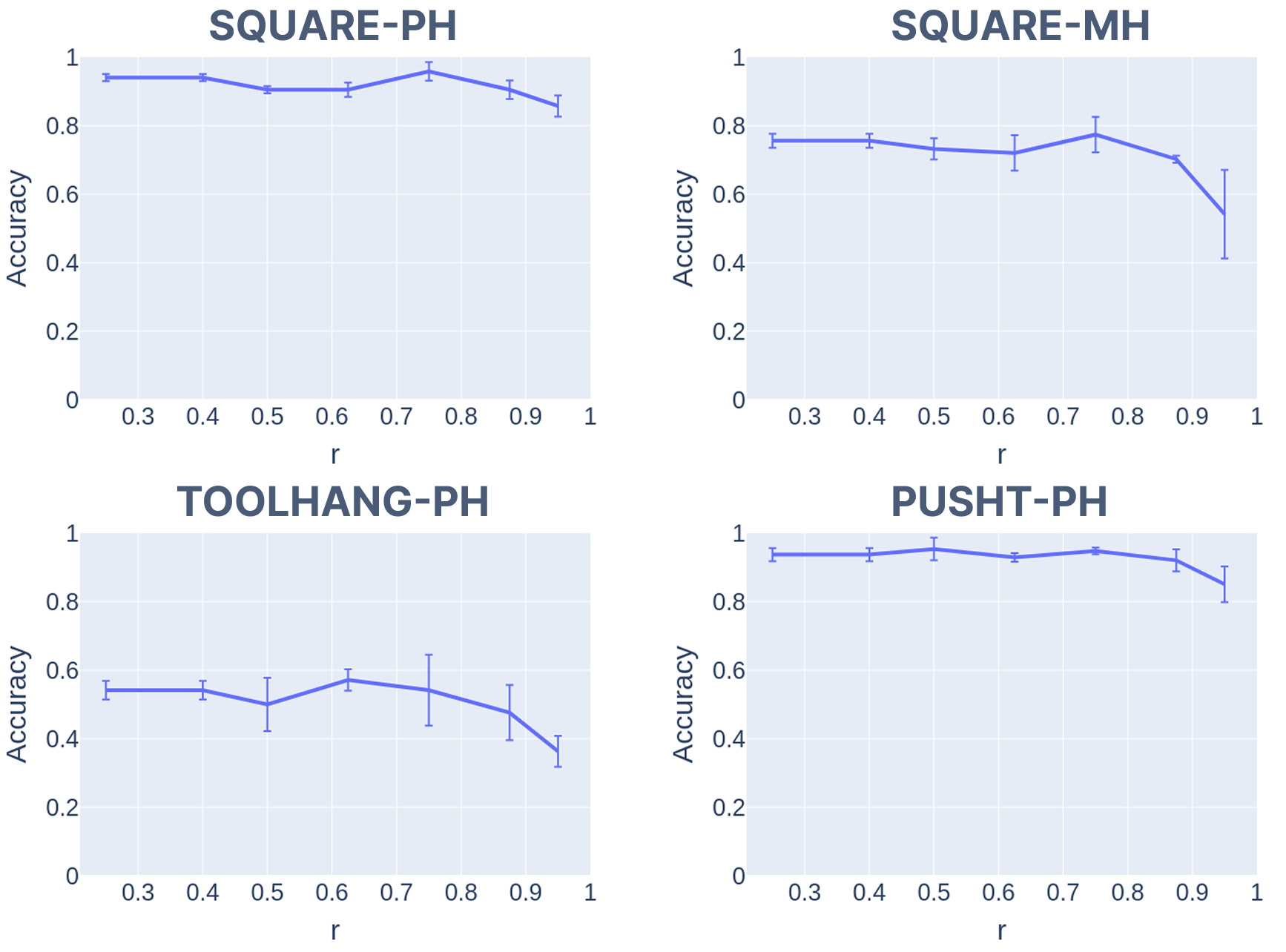}
\end{center}
\caption{\textbf{Accuracy and hyperparameter $r$:} The effect of hyperparameter $r$ on accuracy of the EDM model is shown. It can be seen that the larger the $r$, the faster the generation speed increases, but the accuracy tends to decrease.}
\label{fig:edm_hyper}
\end{figure}

Considering the optimal choice of the hyperparameter $r$, we measure the accuracy when changing the hyperparameter. In image-based diffusion models, there is a trade-off between faithfulness to the input and realism. If the number of steps to denoise with respect to the input is small, realistic samples cannot be generated, and conversely, if the number of denoise steps with respect to the input is large, the faithfulness to the output to be obtained is reduced. Therefore, as $r$ approaches 1, the sampling process becomes faster, but the accuracy is likely to decrease.
We measure changes in $r$ and accuracy using the EDM model with the results shown in \autoref{fig:edm_hyper}.

The graphs show that for each task, performance tends to be stable until $r = 0.75$ where it starts to decrease to a greater extent. Calculating the average recovery rate for benchmark tasks, we found that it was 
103.36\% when $r=0.75$ and
95.72\% when $r=0.875$, 
but dropped to 81.86\% when $r = 0.95$ as seen in \autoref{tab:recovery-rate}. The method offers a $\times 4$ speed improvement while maintaining accuracy, where past this point it becomes more of a trade-off. Therefore, consideration has to be given when selecting the optimal hyperparameter value depending on what metric is more important for the end task.

\section{Conclusion}
In this study, we propose RAGDP, a method for speeding up DP without additional training, by making use of a database from training data to reduce the number of sampling points by using nearby samples as neighborhood points during inference. We provide implementations of RAGDP for both VP-SDE and VE-SDE of Diffusion Models.

We found that RAGDP is effective in improving the trade-off between accuracy and sampling speed when utilized with existing efficient sampling methods like DPM++. 

We also compared CP and RAGDP, where RAGDP-VE outperformed CP in preserving the initial accuracy despite offering the same acceleration capacity. More specifically, for simple tasks or tasks with a small number of action dimensions, it is appropriate to choose between CP and RAGDP based on a balance between accuracy and training cost. For complex tasks, it might be preferable to use RAGDP-VE due to working without additional training, while the training demand increases with task complexity for CP.

While RAGDP can improve the performance of trained DP, there are still areas open for improvement and future work. 
One limitation of the work is that it makes use of demonstration data to make its leaps, which means that the quality and quantity of the data may affect the performance. In our case, Square-MH has low-quality trajectories mixed in the dataset, but RAGDP still performs well, which displays some resistance to low-quality data. Another limitation is the assumption that you have access to the training data, which might not always be available despite efforts being made to open-source datasets along with their models. Furthermore, currently the leap ratio $r$ has to be chosen manually. If the leap ratio can be calculated using a heuristic algorithm, the decision can be dynamically made based on the trade-off between speed and accuracy. Overall, RAGDP offers an off-the-shelf method that can reliably improve the accuracy of DP without additional training.



\printbibliography

\end{document}